\documentclass[11pt,letterpaper]{article}
\usepackage{emnlp2017}
\usepackage{times}
\usepackage{latexsym}
\usepackage{CJK}
\usepackage{url}
\usepackage{multirow}
\usepackage[english]{babel}
\usepackage{natbib}
\usepackage{amsmath}
\usepackage{amssymb}
\usepackage{graphicx}
\usepackage{arydshln}
\usepackage{balance}
\DeclareMathOperator*{\argmax}{argmax}

\emnlpfinalcopy

\def\emnlppaperid{***}

\title{Translating Phrases in Neural Machine Translation}

\author{Xing Wang$^\dag$ { } Zhaopeng Tu$^\ddag$ { } Deyi Xiong$^\dag$\thanks{ \ \ Corresponding author}   { } Min Zhang$^\dag$ \\
 $^\dag${Soochow University, Suzhou, China}\\
  {\tt xingwsuda@gmail.com, \{dyxiong, minzhang\}@suda.edu.cn}\\
  $^\ddag$ {Tencent AI Lab, Shenzhen, China}\\
  {\tt tuzhaopeng@gmail.com} }

\date{}

\begin{document}
\begin{CJK}{UTF8}{gkai}
\maketitle
\begin{abstract}
Phrases play an important role in natural language understanding and machine translation \cite{sag2002multiword, villavicencio2005introduction}. However, it is difficult to integrate them into current neural machine translation (NMT) which reads and generates sentences word by word. In this work, we propose a method to translate phrases in NMT by integrating a phrase memory storing target phrases from a phrase-based statistical machine translation (SMT) system into the encoder-decoder architecture of NMT. At each decoding step, the phrase memory is first re-written by the SMT model, which dynamically generates relevant target phrases with contextual information provided by the NMT model. Then the proposed model reads the phrase memory to make probability estimations for all phrases in the phrase memory. If phrase generation is carried on, the NMT decoder selects an appropriate phrase from the memory to perform phrase translation and updates its decoding state by consuming the words in the selected phrase. Otherwise, the NMT decoder generates a word from the vocabulary as the general NMT decoder does. Experiment results on the Chinese$\rightarrow$English translation show that the proposed model achieves significant improvements over the baseline on various test sets.
\end{abstract}

\section{Introduction}
Neural machine translation (NMT) has been receiving increasing attention due to its impressive translation performance \cite{kalchbrenner2013recurrent,cho2014learning,sutskever2014sequence,bahdanau2014neural,wu2016google}. Significantly different from conventional statistical machine translation (SMT) \cite{brown1993mathematics,koehn2003statistical,chiang2005hierarchical}, NMT adopts a big neural network to perform the entire translation process in one shot, for which an encoder-decoder architecture is widely used. Specifically, the encoder encodes a source sentence into a continuous vector representation, then the decoder uses the continuous vector representation to generate the corresponding target translation word by word.

The word-by-word generation philosophy  in NMT makes it difficult to translate multi-word phrases. Phrases, especially multi-word expressions, are crucial for natural language understanding and machine translation \cite{sag2002multiword, villavicencio2005introduction} as the meaning of a phrase cannot  be always deducible from the meanings of its individual words or parts. Unfortunately current NMT is essentially a word-based or character-based \cite{chung-cho-bengio:2016:P16-1,costajussa-fonollosa:2016:P16-2,luong2016achieving} translation system where phrases are not considered as translation units. In contrast, phrases are much better than words  as translation units in SMT and have made a significant advance in translation quality. Therefore, a natural question arises: Can we translate phrases in NMT?

Recently, there have been some attempts on multi-word phrase generation  in NMT \cite{stahlberg2016syntactically,zhang2016bridging}. However these efforts constrain NMT to generate either syntactic phrases or domain phrases in the word-by-word generation framework. To explore the phrase generation in NMT beyond the word-by-word generation framework, we propose a novel architecture that integrates a phrase-based SMT model into NMT. Specifically, we add an auxiliary phrase memory to store target phrases in symbolic form. At each decoding step, guided by the decoding information from the NMT decoder, the SMT model dynamically generates relevant target phrase translations and writes them to the memory. Then the NMT decoder scores phrases in the phrase memory and selects a proper phrase or word with the highest probability. If the phrase generation is carried out, the NMT decoder generates a multi-word phrase and updates its decoding state by consuming the words in the selected phrase.

Furthermore, in order to enhance the ability of the NMT decoder to effectively select appropriate target phrases, we modify the encoder of NMT to make it fit for exploring structural information of source sentences. Particularly, we integrate syntactic chunk information into the NMT encoder, to enrich the source-side representation.
%Similar to the attention-based NMT model, the proposed model can also be trained in an end-to-end manner.
We validate our proposed model on the Chinese$\rightarrow$English translation task. Experiment results show that the proposed model significantly outperforms the conventional attention-based NMT by 1.07 BLEU points on multiple NIST test sets.

The rest of this paper is organized as follows. Section 2 briefly introduces the attention-based NMT as background knowledge. Section 3 presents our proposed model which incorporates the phrase memory into the NMT encoder-decoder architecture, as well as the reading and writing procedures of the phrase memory. Section 4 presents our experiments on the Chinese$\rightarrow$English translation task and reports the experiment results. Finally we discuss related work in Section 5 and conclude the paper in Section 6.
%and the phrase-based SMT

\section{Background}
%In this section, we briefly describe the attention-based NMT \cite{bahdanau2014neural} and the phrase-based SMT.
%\subsection{Attention-based Neural Machine Translation}
Neural machine translation often adopts the encoder-decoder architecture with recurrent neural networks (RNN) to model the translation process. The bidirectional RNN encoder which consists of a forward RNN and a backward RNN reads a source sentence $\mathbf{x} = x_{1}, x_{2}, ... ,  x_{T_{x}}$ and transforms it into word annotations of the entire source sentence $\mathbf{h} = h_{1}, h_{2}, ..., h_{T_{x}}$. The decoder uses the annotations to emit a target sentence $\mathbf{y} = y_{1}, y_{2}, ..., y_{T_{y}}$ in a word-by-word manner.

In the training phase, given a parallel sentence $(\mathbf{x}, \mathbf{y})$, NMT models the conditional probability as follows,
\begin{equation}
P(\mathbf{y} | \mathbf{x}) = \prod_{i=1}^{T_{y}}{P(y_{i}|\mathbf{y_{<i}}, \mathbf{x})}
\end{equation}
where $y_{i}$ is the target word emitted by the decoder at step $i$ and $\mathbf{y_{<i}} = y_{1}, y_{2}, ..., y_{i-1}$. The conditional probability $P(y_{i}| \mathbf{y_{<i}}, \mathbf{x})$ is computed as
\begin{equation}
P(y_{i}|\mathbf{y_{<i}}, \mathbf{x}) =  softmax(f(s_{i}, y_{i-1}, c_{i}))
\label{eqn-word-prob}
\end{equation}
where $f(\cdot)$ is a non-linear function and $s_{i}$ is the hidden state of the decoder at step $i$:
\begin{equation}
s_{i} = g(s_{i-1}, y_{i-1}, c_{i})
\label{eqn-update}
\end{equation}
where $g(\cdot)$ is a non-linear function. Here we adopt Gated Recurrent Unit \cite{cho2014learning} as the recurrent unit for the encoder and decoder. $c_i$ is the context vector, computed as a weighted sum of the annotations $\mathbf{h}$:
\begin{equation}
c_{i} = \sum_{j=1}^{T_{x}}\alpha_{t,j} h_{j}
\label{eqn-context}
\end{equation}
where $h_{j}$ is the annotation of source word $x_{j}$ and its weight $\alpha_{t,j}$ is computed by the attention model.

We train the attention-based NMT model by maximizing the log-likelihood:
\begin{equation}
C(\theta) =\sum_{n=1}^{N} \sum_{i=1}^{T_{y}} \log P(y_{i}^{n}  | \mathbf{y}_{<i}^{n}, \mathbf{x}^{n})
\end{equation}
given the training data with $N$ bilingual sentences \cite{cho2015natural}.

In the testing phase, given a source sentence $\mathbf{x}$, we use beam search strategy to search a target sentence $\hat{\mathbf{y}}$ that approximately maximizes the
conditional probability $P(\mathbf{y} | \mathbf{x})$
\begin{equation}
\hat{\mathbf{y}} = \mathop{\argmax}_{\mathbf{y}}{P(\mathbf{y} | \mathbf{x})}
\end{equation}

\begin{figure}[t]
\centerline{\includegraphics[width=220pt]{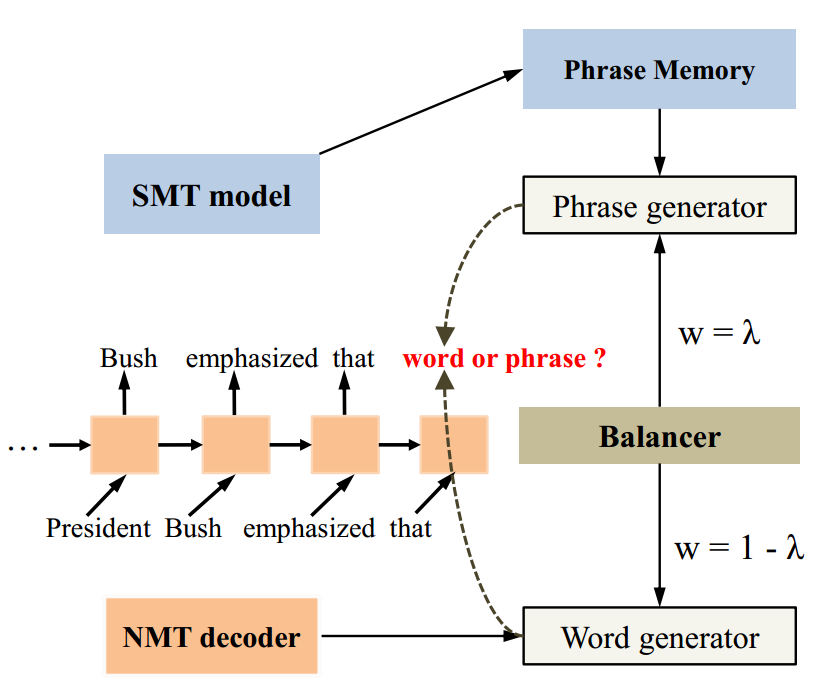}}
   % \vspace{-2em}
\caption{Architecture of the NMT decoder with the phrase memory. The NMT decoder performs phrase generation using the balancer and the phrase memory. }\label{figure:1}
\end{figure}

\section{Approach}

In this section, we introduce the proposed model which incorporates a phrase memory into the encoder-decoder architecture of NMT. Inspired by the recent work on attaching an external structure to the encoder-decoder architecture \cite{gulcehre-EtAl:2016:P16-1,gu2016incorporating,tang2016neural,wang2016neural}, we adopt a similar approach to incorporate the phrase memory into NMT.

\subsection{Framework}

Figure~\ref{figure:1} shows an example. Given the generated words ``{\em President Bush emphasized that}'', the model generates the next fragment either from a word generation mode or a phrase generation mode.
If the model selects the word generation mode, it generates a word by the NMT decoder as in the standard NMT framework.
Otherwise, it generates a multi-word phrase by enquiring a phrase memory, which is written by an SMT decoder based on the dynamic decoding information from the NMT model for each step.
The trade-off between word generation mode and phrase generation mode is balanced by a weight $\lambda$, which is produced by a neural network based {\em balancer}.

Formally, a generated translation ${\bf y}=\{y_1, y_2, \dots, y_{T_{y}}\}$ consists of two sets of fragments: words generated by NMT decoder ${\bf w}=\{w_1, w_2, \dots, w_K\}$ and phrases generated from the phrase memory ${\bf p}=\{p_1, p_2, \dots, p_L\}$ . The probability of generating ${\bf y}$ is calculated by
\begin{eqnarray}
P(\mathbf{y} | \mathbf{x}) = \prod_{w_k \in {\bf w}} (1-\lambda_{t(w_k)}) P_{word}(w_k) \nonumber \\
					 \times \prod_{p_l \in {\bf p}} \lambda_{t(p_l)} P_{phrase}(p_l)
\label{eqn-probability}
\end{eqnarray}
where $P_{word}(w_k)$ is the probability of generating the word $w_k$ (see Equation~\ref{eqn-word-prob}), $P_{phrase}(p_l)$ is that of generating the phrase $p_l$ which will be described in Section~\ref{sec-memory},  and $t(\cdot)$ is the decoding step to generate the corresponding fragment.

 The balancing weight $\lambda$ is produced by the {\em balancer} -- a  multi-layer network. The balancer network takes as input the decoding information, including the context vector $c_i$, the previous decoding state $s_{i-1}$ and the previous generated word $y_{i-1}$:
\begin{equation}
\lambda_i = \sigma(f_b(s_i, y_{i-1}, c_i))
\label{eqn-lambda}
\end{equation}
where $\sigma(\cdot)$ is a sigmoid function and $f_b(\cdot)$ is the activation function.
Intuitively, the weight $\lambda$ can be treated as the estimated importance of the phrase to be generated. We expect $\lambda$ to be high if the phrase is appropriate at the current decoding step.

\paragraph{Well-Formed Phrases}

We employ a source-side chunker to chunk the source sentence, and only phrases that corresponds to a source chunk are used in our model.
We restrict ourselves to the well-formed chunk phrases based on the following considerations:
(1) In order to take advantage of dynamic programming, we restrict ourselves to non-overlap phrases.\footnote{Overlapped phrases may result in a high dimensionality in translation hypothesis representation and make it hard to employ shared fragments for efficient dynamic programming.}
(2) We explicitly utilize the boundary information of the source-side chunk phrases, to better guide the proposed model to adopt a target phrase at an appropriate decoding step.
(3) We enable the model to exploit the syntactic categories of chunk phrases to enhance the proposed model with its selection preference for special target phrases. With these information, we enrich the context vector $c_i$ to enable the proposed model to make better decisions, as described below.

Following the commonly-used strategy in sequence tagging tasks~\cite{xue-shen:2003:SIGHAN}, we allow the words in a phrase to share the same chunk tag and introduce a special tag for the beginning word. For example, the phrase `` 信息 安全 (information security)'' is tagged as a noun phrase ``NP'', and the tag sequence should be ``NP\_B NP''. Partially motivated by the work on integrating linguistic features into NMT \cite{sennrich-haddow:2016:WMT}, we represent the encoder input as the combination of word embeddings and chunking tag embeddings, instead of word embeddings alone in the conventional NMT. The new input is formulated as follows:
\begin{equation}
[E^{w}x_i, E^{t}t_i]
\end{equation}
where $E^{w} \in \mathbb{R}^{dw \times |V^{NMT}|}$ is a word embedding matrix and $dw$ is the word embedding dimensionality, $E^{t} \in \mathbb{R}^{dt \times |V^{TAG}|}$ is a tag embedding matrix and $dt$ is the tag embedding dimensionality. $[\cdot]$ is the vector concatenation operation.

\subsection{Phrase Memory}
\label{sec-memory}

The phrase memory stores relevant target phrases provided by an SMT model, which is trained on the same bilingual corpora.
At each decoding step, the memory is firstly erased and  re-written by the SMT model, the decoding of which is based on the translation information provided by the NMT model.
Then, the proposed model enquires phrases along with their probabilities $P_{phrase}$ from the memory.

\paragraph{Writing to Phrase Memory}
Given a partial translation ${\bf y}_{<i} = \{y_1, y_2, \dots, y_{t-1}\}$ generated from NMT, the SMT model picks potential phrases extracted from the translation table. The phrases are scored with multiple SMT features, including the language model score, the translation probabilities, the reordering score, and so on. Specially, the reordering score depends on alignment information between source and target words, which is derived from attention distribution produced by the NMT model~\cite{wang2016neural}. SMT coverage vector in ~\cite{wang2016neural} is also introduced to avoid repeat phrasal recommendations. In our work, the potential phrase is phrase with high SMT score which is defined as following:
\begin{equation}
SMT_{score}(p_{l}  | \mathbf{y_{<t}}, \mathbf{x}) =  \sum_{m=1}^{M}w_{m}h_{m}(p_{l}, x(p_l))
\label{eqn-SMT-score}
\end{equation}
where $p_{l}$ is a target phrase and $x(p_l)$ is its corresponding source span. $h_{m}(p_{l}, x(p_l))$ is a SMT feature function and $w_{m}$ is its weight. The feature weights can be tuned by the minimum error rate training (MERT) algorithm \cite{och2003minimum}.

This leads to a better interaction between SMT and NMT models. It should be emphasized that our memory is dynamically updated at each decoding step based on the decoding history from both SMT and NMT models.

The proposed model is very flexible, where the phrase memory can be either fully dynamically generated by an SMT model or directly extracted from a bilingual dictionary, or any other bilingual resources storing idiomatic translations or bilingual multi-word expressions, which may lead to a further improvement.
\footnote{Bilingual resources can be utilized in two ways: First, we can store the bilingual resources in a static memory and keep all items available to NMT in the whole decoding period. Second, we can integrate the bilingual resources into SMT and then dynamically feed them into the phrase memory.}

\paragraph{Reading Phrase Memory}

When phrases are read from the memory, they are rescored by a neural network based score function. The score function takes as input the phrase itself and  decoding information from NMT ($i=t(p_l)$ denotes the current decoding step):
\begin{eqnarray}
score_{phrase}(p_l) = g_{s}\big(e(p_l), s_i, y_{i-1}, c_i\big)
\label{eqn-score-memory}
\end{eqnarray}
where $g_{s}(\cdot)$ is either an identity or a non-linear function.  $e(p_l)$ is the representation of phrase $p_l$, which is modeled by a recurrent neural networks. Again, $s_i$ is the decoder state, $y_{i-1}$ is the lastly generated word, and $c_i$ is the context vector.
The scores are normalized for all phrases in the phrase memory, and the probability for phrase $p_l$ is calculated as
\begin{eqnarray}
P_{phrase}(p_l) = softmax(score_{phrase}(p_l))
\end{eqnarray}
The probability calculation is controlled with parameters, which are trained together with the parameters from the NMT model.

\subsection{Training}
%The proposed model can be trained in an end-to-end fashion.
Formally, we train both the default parameters of standard NMT and the new parameters associated with phrase generation on a set of training examples $\{\left[{\bf x}^n, {\bf y}^n\right]\}_{n=1}^{N}$:
\begin{eqnarray}
C(\theta)        =  \sum_{n=1}^{N} \log P(\mathbf{y}^{n} | \mathbf{x}^{n})
\end{eqnarray}
where
$P(\mathbf{y}^{n} | \mathbf{x}^{n})$ is defined in Equation~\ref{eqn-probability}.
Ideally, the trained model is expected to produce a higher balance weight $\lambda$ and phrase probability $P_{phrase}$ when a phrase is selected from the memory, and lower scores in other cases.

\subsection{Decoding}
\label{sec-decoding}
During testing, the NMT decoder generates a target sentence which consists of a mixture of words and phrases.
Due to the different granularities of words and phrases, we design a variant of beam search strategy:
At decoding step $i$, we first compute $P_{phrase}$ for all phrases in the phrase memory and $P_{word}$ for all words in NMT vocabulary.
Then the balancer outputs a balancing weight $\lambda_i$, which is used to scale the phrase and word probabilities :  $\lambda_i \times P_{phrase}$ and $ (1 - \lambda_i) \times P_{word}$.
Now outputs are normalized probabilities on the concatenation of phrase memory and the general NMT vocabulary. At last, the NMT decoder generates a proper phrase or word of the highest probability.

If a target phrase in the phrase memory has the highest probability, the decoder generates the target phrase to complete the multi-word phrase generation process, and updates its decoding state by consuming the words in the selected phrase as described in Equation~\ref{eqn-update}. All translation hypotheses are placed in the corresponding beams according to the number of generated target words.

\section{Experiments}
In this section, we evaluated the effectiveness of our model on the Chinese$\rightarrow$English machine translation task.
The training corpora consisted of about 1.25 million sentence pairs\footnote{The corpus includes LDC2002E18, LDC2003E07, LDC2003E14, Hansards portion of LDC2004T07, LDC2004T08 and LDC2005T06.} with 27.9 million Chinese words and 34.5 million English words respectively.
We used NIST 2006 (NIST06) dataset as development set, and NIST 2004 (NIST04), 2005 (NIST05) and 2008 (NIST08) datasets as test sets. We report experiment results with case-insensitive BLEU score\footnote{ftp://jaguar.ncsl.nist.gov/mt/resources/mteval-v11b.pl}.

\begin{table*}[t]
\centering
\begin{tabular}{l| c c c c}
\hline
SYSTEM & NIST04 & NIST05 & NIST08 & Avg\\
\hline
\hline
Moses                                                      & 34.74           & 31.99           & 23.69           &  30.14 \\
RNNSearch                                                  & 37.80           & 34.70           & 24.93           &  32.48 \\
\hline
\ \ \  +memory                                              & 38.21           & 35.15           & 25.48$\dag$           & 32.95 \\
\ \ \  +memory +chunking tag                                & 38.83$\ddag$    & 35.72$\ddag$    & 26.09$\ddag$          & 33.55 \\
\hline
\end{tabular}
\caption{Main experiment results on the NIST Chinese-English translation task. BLEU scores in the table are case insensitive. Moses and RNNSearch are SMT and NMT baseline system respectively. ``$\dag$'': significantly better than RNNSearch ($p < 0.05$); ``$\ddag$'': significantly better than RNNSearch ($p < 0.01$).}\label{table:main}
\end{table*}

%Main experiment results on the NIST Chinese$\rightarrow$English translation task. The BLEU scores are case-insensitive.
%
%``$\dag$'': significantly better than RNNSearch ($p < 0.05$); ``$\ddag$'': significantly better than RNNSearch ($p < 0.01$).

We compared our proposed model with two state-of-the-art systems:
\begin{itemize}
\item[*]  \textbf{Moses}: a state-of-the-art phrase-based SMT system \cite{koehn-EtAl:2007:PosterDemo} with its default settings, where  feature function weights are tuned by the minimum error rate training (MERT) algorithm \cite{och2003minimum}.
\item[*]  \textbf{RNNSearch}:  an in-house implementation of the attention-based NMT system \cite{bahdanau2014neural} with its default settings. %We also adopt the dropout technique to strengthen our system.
\end{itemize}

For Moses, we used the full bilingual training data to train the phrase-based SMT model and the target portion of the bilingual training data to train a 4-gram language model using  KenLM\footnote{https://kheafield.com/code/kenlm/}. We ran Giza++ on the training data in both Chinese-to-English and English-to-Chinese directions and applied the ``grow-diag-final'' refinement rule \cite{koehn2003statistical} to obtain word alignments. The maximum phrase length is set to 7.

For RNNSearch, we generally followed settings in the previous work \cite{bahdanau2014neural,tu2016context,tu2016neural}. We only kept a shortlist of the most frequent 30,000 words in Chinese and English, covering approximately 97.7\% and 99.3\% of the data in the two languages respectively.
%Words that are not in the shortlist were mapped to a special token UNK.
We constrained our source and target sequences to have a maximum length of 50 words in the training data.
% and used the constrained parallel data to train the model.
The size of embedding layer of both sides was set to 620 and the size of hidden layer was set to 1000. We used a minibatch stochastic gradient descent (SGD) algorithm of size 80 together with Adadelta \cite{zeiler2012adadelta} to train the NMT models. The decay rates $\rho$ and $\epsilon$ were set as $0.95$ and $10^{-6}$. We clipped the gradient norm to 1.0 \cite{pascanu2013difficulty}. We also adopted the dropout technique. Dropout was applied only on the output layer and the dropout rate was set to 0.5. We used a simple beam search decoder with beam size 10 to find the most likely translation.

For the proposed model, we used a Chinese chunker\footnote{http://www.niuparser.com/} \cite{zhu-EtAl:2015:ACL-IJCNLP-2015-System-Demonstrations} to chunk the source-side Chinese sentences. 13 chunking tags appeared in our chunked sentences and the size of chunking tag embedding was set to 10. We used the trained phrase-based SMT to translate the source-side chunks. The top 5 translations according to their translation scores (Equation~\ref{eqn-SMT-score}) were kept and among them multi-word phrases were used as phrasal recommendations  for each source chunk phrase. For a source-side chunk phrase, if there exists phrasal recommendations from SMT, the output chunk tag was used as its chunking tag feature as described in Section 3.1. Otherwise, the words in the chunk were treated as general words by being tagged with the default tag.
In the phrase memory, we only keep the top 7 target translations with highest SMT scores at each decoding step. We used a forward neural network with two hidden layers for both the balancer (Equation~\ref{eqn-lambda}) and the scoring function (Equation~\ref{eqn-score-memory}). The numbers of units in the hidden layers were set to 2000 and 500 respectively. We used a backward RNN encoder to learn the phrase representations of target phrases in the phrase memory.

\subsection{Main Results}
Table~\ref{table:main} reports main results of different models measured in terms of BLEU score. We observe that our implementation of RNNSearch outperforms Moses by 2.34 BLEU points. (\emph{+memory}) which is the proposed model with the phrase memory obtains an improvement of  0.47 BLEU points over the baseline RNNSearch. With the source-side chunking tag feature, (\emph{+memory+chunking tag}) outperforms the baseline RNNSearch by 1.07 BLEU points, showing the effectiveness of chunking syntactic categories on the selection of appropriate target phrases.
From here on, we use ``{\em +memory+chunking tag}'' as the default setting in the following experiments if not otherwise stated.

\begin{table}[t]
\centering
\begin{tabular}{l| c c c c}
\hline
 & NIST04 & NIST05 & NIST08 \\
\hline
+memory                              & 34.3\%           & 29.4\%           & 22.2\%   \\
\ +chunking tag                  & 66.4\%           & 63.1\%           & 58.4\%    \\
\hline
\end{tabular}
\caption{Percentages of sentences that contain phrases generated by the proposed model. }\label{table:sentence}
\end{table}

\paragraph{Number of Sentences Affected by Generated Phrases}
We also check the number of translations that contain phrases generated by the proposed model, as shown in Table~\ref{table:sentence}.
As seen, a large portion of translations take the recommended phrases, and the number increases when the chunking tag feature is used.\footnote{The numbers on NIST08 are relatively lower since part of the test set contains sentences from Web forums, which contain less multi-word expressions.} Considering BLEU scores reported in Table~\ref{table:main}, we believe that the chunking tag feature benefits the proposed model on its phrase generation.

\subsection{Analysis on Generated Phrases}

\begin{table}[h]
\centering
\begin{tabular}{c| rr|rr}
\hline
%Category  & Total	&	Correct \\
%\hline
%NP	&	81.0\%	&	38.9\%\\
%VP	&	8.0\%	&	1.7\%\\
%QP	&	10.8\%	&	4.3\%\\
%Others	&	0.2\%	&	0\%\\
%\hline
%Sum	&	100\%	&	44.9\%\\
\multirow{2}{*}{Type}  &   \multicolumn{2}{c|}{All}	&	\multicolumn{2}{c}{New}\\
\cline{2-5}
					&	Total	&	Correct		&	Total	&	Correct\\
\hline
NP	&	81.0\%	&	38.7\%	&	46.0\%	&	11.5\%\\
VP	&	8.0\%	&	1.7\%	&	6.5\%	&	0.8\%\\
QP	&	10.8\%	&	4.1\%	&	6.2\%	&	0.9\%\\
Others	&	0.2\%	&	0\%	&	0.2\%	&	0\%\\
\hline
Sum	&	100\%	&	44.5\%	&	58.9\%	&	13.2\%\\
\hline
\end{tabular}
\caption{Percentages of phrase categories to the total number of generated ones. ``All'' denotes all generated phrases, and ``New'' means new phrases that cannot be found in translations generated by the baseline system.
``Total'' is the total number of generated phrases and ``Correct'' denotes the fully correct ones.}\label{table:category}
\end{table}

\paragraph{Syntactic Categories of Generated Phrases} We first investigate which category of phrases is more likely to be selected by the proposed approach. There are some phrases, such as noun phrases (NPs, e.g., ``national laboratory'' and ``vietnam airlines'') and quantifier phrases (QPs, e.g., ``15 seconds'' and ``two weeks'') , that we expect to be favored by our approach.
Statistics shown in Table~\ref{table:category} confirm our hypothesis.
Let's first concern all generated phrases (i.e., column ``All''): most selected phrases are noun phrases (81.0\%) and quantifier phrases (10.8\%).
%This is intuitive, since multi-word expressions (e.g., idioms and dates) generally belong to the two categories, which is exactly the motivation of the proposed model.
Among them, 44.5\% percent of them are fully correct\footnote{Fully correct means that the generated phrases can be retrieved in corresponding references as a whole unit.}. Specifically, NPs have relative higher generation accuracy (i.e., $47.8\%=38.7\% / 81.0\%$) while VPs have lower accuracy (i.e., $21.2\% = 1.7\% / 8.0\%$). By looking into the wrong cases, we found most errors are related to verb tense, which is the drawback of SMT models.

Concerning the newly introduced phrases that cannot be found in baseline translations (i.e., column ``New''),
13.2\% of generated phrases are both new and fully correct, which contribute most to the performance improvement. We can also find that most newly introduced verb phrases and quantifier phrases are not correct, the patterns of which can be well learned by word-based NMT models.

\begin{table}[h]
\centering
\begin{tabular}{c|rr|rr}
%\hline
%Words  & Total	&	Correct\\
%\hline
%2	&	66.2\%	&	34.4\%\\
%3	&	20.7\%	&	8.2\%\\
%4	&	7.4\%	&	1.8\%\\
%$\geqslant$5	&	5.7\%	&	0.6\%\\
%\hline
\hline
\multirow{2}{*}{Words}  &   \multicolumn{2}{c|}{All}	&	\multicolumn{2}{c}{New}\\
\cline{2-5}
					&	Total	&	Correct		&	Total	&	Correct\\
\hline
2	&	66.2\%	&	33.6\%	&	34.9\%	&	9.1\%\\
3	&	20.7\%	&	8.4\%	&	13.4\%	&	3.2\%\\
4	&	7.4\%	&	1.9\%	&	5.4\%	&	0.6\%\\
$\geqslant$5	&	5.7\%	&	0.6\%	&	5.2\%	&	0.3\%\\
\hline
\end{tabular}
\caption{Percentages of phrases with different word counts to the total number of generated ones.}\label{table:order}
\end{table}

\paragraph{Number of Words in Generated Phrases}
Table~\ref{table:order} lists the distribution of generated phrases based on the number of inside words.
As seen, most generated phrases are short phrases (e.g., 2-gram and 3-gram phrases), which also contribute most to the new and fully correct phrases (i.e., $12.3\%=9.1\%+3.2\%$). Focusing on long phrases (e.g., order$\geqslant4$), most of them are newly introduced ($10.6\%$ out of $13.1\%$). Unfortunately, only a few portion of these phrases are fully correct, since long phrases have higher chance to contain one or two unmatched words.

\begin{table}[h]
\centering
\begin{tabular}{l| c }
\hline
SYSTEM & Test\\
\hline
\hline
\ \ \  +memory                                              & 32.95 \\
\ \ \  +memory +NULL                                & 31.63 \\
\hline
\ \ \  +memory +chunking tag                     & 33.55 \\
\ \ \  +memory +chunking tag +NULL      & 30.81 \\
\hline
\end{tabular}
\caption{Additional experiment results on the translation task to directly measure the improvement obtained by the phrase generation. ``+NULL'' denotes that we replace the generated target phrases with a special symbol “NULL” in test sets. BLEU scores in the table are case insensitive.}\label{table:additional}
\end{table}

\paragraph{Effect of Generated Phrases on Translation Performance}
Note that the proposed model benefits not only from fully matched phrases, but also from partially matched phrases. For example, the baseline system translates `` 国家 航空 暨 太空 总署'' in a word-by-word manner and outputs ``state aviation and space department''. The generated phrase provided by SMT is ``national aviation and space administration", but the only correct reference is ``national aeronautics and space administration". The generated phrase is not fully correct but still useful.

To directly measure the improvement obtained by the phrase generation, we replace the generated target phrases with a special symbol ``NULL'' in test sets.
As shown in Table~\ref{table:additional}, when deleting the generated target phrases, (``\emph{+memory+chunking tag}") and (``\emph{+memory}") translation performances decrease by 2.74 BLEU points and 1.32 BLEU points respectively. Moreover, translation performances on NIST08 decrease less than those on NIST04 and NIST05 in both settings. The reason is that NIST08 which contains sentences from web data has little influence on generating target phrases which are provided from a different domain \footnote{The parallel training data are mainly from news domain.}.  The overall results demonstrate that neural machine translation benefits from phrase translation.

\subsection{Effect of Balancer}

\begin{table}[h]
\centering
\begin{tabular}{l| c }
\hline
Weight & Test\\
\hline
Dynamic                                              & 33.55 \\
Constant ($\lambda=0.1$)                                & 31.35 \\
\hline
\end{tabular}
\caption{Translation performance with a variety of balancing weight strategies. ``Dynamic'' is the proposed approach and ``Constant ($\lambda=0.1$)'' denotes fixing the balancing weight to 0.1. BLEU scores in the table are case insensitive.}\label{table:lambda}
\end{table}

The balancer which is used to coordinate the phrase generation and word generation is very crucial for the proposed model. We conducted an additional experiment to validate the effectiveness of the neural network based balancer.
%To eliminate the influence of chunking tag features
We use the setting ``+memory +chunking tag'' as baseline system to conduct the experiments. In this experiment, we fixed the balancing weight $\lambda$ (Equation~\ref{eqn-lambda}) to 0.1 during training and testing and report the results. As shown in Table~\ref{table:lambda}, we find that using the fixed value for the balancing weight (\emph{Constant ($\lambda=0.1$) }) decreases the translation performance sharply. This demonstrates that the neural network based balancer is an essential component for the proposed model.

\subsection{Comparison to Word-Level Recommendations and Discussions}
\begin{table}[h]
\centering
\begin{tabular}{l| c }
\hline
SYSTEM & Test\\
\hline
\hline
\ \ \  +word level recommendation                              & 33.27 \\
\ \ \  +memory +chunking tag                     & 33.55 \\
\hline
\end{tabular}
\caption{Experiment results on the translation task. ``+word level recommendation'' is the proposed model in \cite{wang2016neural}. BLEU scores in the table are case insensitive.}\label{table:wang}
\end{table}

Our approach is related to our previous work \cite{wang2016neural} which integrates the SMT word-level knowledge into NMT. To make a comparison, we conducted experiments followed settings in \cite{wang2016neural}. The comparison results are reported in Table~\ref{table:wang}. We find that our approach is marginally better than the word-level model proposed in \cite{wang2016neural} by 0.28 BLEU points.

In our approach, the SMT model translates source-side chunk phrases using the NMT decoding information. Although we use high-quality target phrases as phrasal recommendations, our approach still suffers from the errors in segmentation and chunking. For example, the target phrase ``laptop computers'' cannot be recommended by the SMT model if the Chinese phrase ``手 提 电脑'' is not chunked as a phrase unit. This is the reason why some sentences do not have corresponding phrasal recommendations (Table~\ref{table:sentence}). Therefore,  our approach can be further enhanced if we can reduce the error propagations from the segmenter or chunker, for example, by using n-best chunk sequences instead of the single best chunk sequence.

Additionally, we also observe that some target phrasal recommendations have been also generated by the baseline system in a word-by-word manner. These phrases, even taken as parts of final translations by the proposed model, do not lead to improvements in terms of BLEU as they have already occurred in translations from the baseline system. For example, the proposed model successfully carries out the phrase generation  mode to generate a target phrase ``guangdong province'' (the translation of Chinese phrase ``广东省'') which has appeared in the baseline system.

As external resources, e.g., bilingual dictionary, which are complementary to the SMT phrasal recommendations, are compatible with the proposed model, we believe that the proposed model will get further improvement by using external resources.

\section{Related work}
Our work is related to the following research topics on NMT:

\paragraph{Generating phrases for NMT}  In these studies, the generated NMT multi-word phrases are either from an SMT model or a bilingual dictionary.  In syntactically guided neural machine translation (SGNMT), the NMT decoder uses phrase translations produced by the hierarchical phrase-based SMT system Hiero, as hard decoding constraints. In this way, syntactic phrases are generated by the NMT decoder \cite{stahlberg2016syntactically}. \citet{zhang2016bridging} use an SMT translation system, which is integrated an additional bilingual dictionary, to synthesize pseudo-parallel sentences and feed the sentences into the training of NMT in order to translate low-frequency words or phrases. \citet{tang2016neural} propose an external phrase memory that stores phrase pairs in symbolic forms for NMT. During decoding, the NMT decoder enquires the phrase memory and properly generates phrase translations. The significant differences between these efforts and ours are 1) that we dynamically generate phrase translations via an SMT model, and 2) that at the same time we modify the encoder to incorporate structural information to enhance the capability of NMT in phrase translation.
%(SGNMT forces the NMT decoder to generate a target word from a Hiero lattice)

\paragraph{Incorporating linguistic information into NMT} NMT is essentially a sequence to sequence mapping network that treats the input/output units, eg., words, subwords \cite{sennrich-haddow-birch:2016:P16-12}, characters \cite{chung-cho-bengio:2016:P16-1, costajussa-fonollosa:2016:P16-2}, as non-linguistic symbols. However, linguistic information can be viewed as the task-specific knowledge, which may be a useful supplementary to the sequence to sequence mapping network. To this end, various kinds of linguistic annotations have been introduced into NMT to improve its translation performance. \citet{sennrich-haddow:2016:WMT} enrich the input units of NMT with various linguistic features, including lemmas, part-of-speech tags, syntactic dependency labels and morphological features.
%They concatenate feature embeddings with word embeddings and feed the concatenated embeddings into the NMT encoder.
\citet{garcia2016factored} propose factored NMT using the morphological and grammatical decomposition of the words (factors) in output units. \citet{eriguchi-hashimoto-tsuruoka:2016:P16-1} explore the phrase structures of input sentences and propose a tree-to-sequence attention model for the vanilla NMT model.  \citet{li2017modeling} propose to linearize source-side parse trees to obtain structural label sequences and explicitly incorporated the structural sequences into NMT, while \citet{aharoni2017towards} propose to incorporate target-side syntactic information into NMT by serializing the target sequences into linearized, lexicalized constituency trees. \citet{zhang-EtAl:2016:COLING3} integrate topic knowledge into NMT for domain/topic adaptation.

\paragraph{Combining NMT and SMT} A variety of approaches have been explored for leveraging the advantages of both NMT and conventional SMT. \citet{he2016improved} integrate SMT features with the NMT model under the log-linear framework in order to help NMT alleviate the limited vocabulary problem \cite{luong2014addressing,chousing} and coverage problem \cite{tu2016modeling}. \citet{arthur2016incorporating} observe that NMT is prone to making mistakes in translating low-frequency content words and therefore attempt at incorporating discrete translation lexicons into the NMT model, to alliterate the imprecise translation problem \cite{wang2016neural}. Motivated by the complementary strengths of syntactical SMT and NMT, different combination schemes of Hiero and NMT have been exploited to form SGNMT \cite{stahlberg2016neural,stahlberg2016syntactically}. \citet{wang2016neural} propose an approach to incorporate the SMT model into attention-based NMT. They combine NMT posteriors with SMT word recommendations through linear interpolation implemented by a gating function which dynamically assigns the weights. \citet{niehues-EtAl:2016:COLING} propose to use SMT to pre-translate the inputs into target translations and employ the target pre-translations as input sequences in NMT.  \citet{zhou2017neural} propose a neural system combination framework to directly combine NMT and SMT outputs. The combination of NMT and SMT has been also introduced in interactive machine translation to improve the system's suggestion quality \cite{wuebker-EtAl:2016:P16-1}. In addition, word alignments from the traditional SMT pipeline are also used to improve the attention mechanism in NMT \cite{cohn-EtAl:2016:N16-1,mi-wang-ittycheriah:2016:EMNLP2016,liu-EtAl:2016:COLING}.

\section{Conclusion}
In this paper, we have presented a novel model to translate source phrases and generate target phrase translations in NMT by integrating the phrase memory into the encoder-decoder architecture. At decoding, the SMT model dynamically generates relevant target phrases with contextual information provided by the NMT model and writes them to the phrase memory. Then the proposed model reads the phrase memory and uses the balancer to make probability estimations for the phrases in the phrase memory. Finally the NMT decoder selects a phrase from the phrase memory or a word from the vocabulary of the highest probability to generate. Experiment results on  Chinese$\rightarrow$English translation have demonstrated that the proposed model can significantly improve the translation performance.

\section*{Acknowledgments}
We would like to thank three anonymous reviewers for their insightful comments, and also acknowledge Zhengdong Lu, Lili Mou for useful discussions. This work was supported by the National Natural Science Foundation of China (Grants No.61525205, 61373095 and 61622209).

\bibliography{emnlp2017}
\bibliographystyle{emnlp_natbib}
\end{CJK}
\end{document}